\documentclass[10pt,conference]{IEEEtran}
\IEEEoverridecommandlockouts
\usepackage{cite}
\usepackage{amsmath,amssymb,amsfonts}
\usepackage{graphicx}
\usepackage{textcomp}
\usepackage{xcolor}
\usepackage[hidelinks]{hyperref}
\usepackage{flushend}
\def\BibTeX{{\rm B\kern-.05em{\sc i\kern-.025em b}\kern-.08em
    T\kern-.1667em\lower.7ex\hbox{E}\kern-.125emX}}

\begin{document}
\title{Skill-Composer: Requirement-Driven Skill Selection from Multiple Marketplaces}

\author{%
\IEEEauthorblockN{Zicheng Chen\textsuperscript{1}, Jia Li\textsuperscript{1,*}, Xiaohong Chen\textsuperscript{2}, and Zhi Jin\textsuperscript{1,3,*}}
\IEEEauthorblockA{\textsuperscript{1}School of Computer Science, Wuhan University, \textsuperscript{2}East China Normal University, \\ \textsuperscript{3}The Key Laboratory of High Confidence Software Technologies, Ministry of Education, Peking University}
\thanks{*Corresponding authors.}
}
\maketitle

\begin{abstract}
Agent skills have become the primary vehicle through which agents reuse specialized capabilities, yet public skill marketplaces provide skills with individual items: a query, whether matched through keyword or embedding similarity, returns a list of single skills ordered by relevance or by popularity. Such an interface cannot answer the question a developer actually faces: which set of skills should be installed to cover a task's complete requirement? The highest-ranked skills are often functionally redundant, while capabilities that are genuinely required remain uncovered. In this paper, we argue that skill acquisition should be reframed from ranking to set selection. We propose Skill-Composer, an agent-native framework that selects a non-redundant skill bundle for the user requirement, which aims to cover all available required capabilities and report any gaps. Specifically, it clarifies user requirements into atomic requirement points, retrieves and merges candidate skills across fragmented marketplaces, decomposes each candidate's capabilities, and progressively discards overlapping ones, so that each retained skill provides different capabilities. This moves the decision of which skills to acquire from the developer to the agent itself. We position this reframing and sketch its workflow.
\end{abstract}


\section{Introduction}
Agent skills package reusable instructions and specialized expertise that an agent can load for a task \cite{registry}. Instead of re-deriving how to profile a program or refactor a module, an agent can load a packaged skill that encodes that expertise. This reuse extends an agent's competence without retraining it and lets practitioners compose solutions over a shared ecosystem. Today the skill ecosystem presents a supply that is at once vast and fragmented: as of November 2026, the SkillsMP platform alone indexes over 4 million SKILL.md files \cite{skillsmp}, the ClawHub marketplace hosts over 85,000 skills \cite{clawhub}, and SkillNet curates over five hundred thousand skills \cite{skillnet}. No single catalog holds the whole supply, and selecting from a library at this scale becomes a bottleneck \cite{toollery}.

However, most existing retrieval modes, whether keyword matching, embedding-based semantic search, or popularity ranking \cite{skillnet}, return a list of single skills ordered by how close each skill is to the query. That answers which skill is most similar to a query, not the question a developer actually faces: which set of skills should be installed so that the task is completely covered? This limitation does not come from retrieval quality; it comes from the decision unit of these interfaces, a single skill, and it surfaces on two levels: whether the retrieved skills cover the full user requirement, and whether candidates are redundant. \textbf{Query relevance is not requirement alignment}: A query is a few keywords, whereas a task carries several subrequirements, some stated and some implicit. The query-relevance score says nothing about which of those requirements a skill actually satisfies. Existing research bears this out: skills that anyone can download from these marketplaces often yield no significant improvement \cite{swe-skills-bench}. \textbf{Candidate skills are mutually redundant}: Marketplaces host many skills with overlapping capabilities. Because each candidate is scored on its own, the overlap goes unnoticed, and the top of the list can fill with near-duplicates while a capability the task requires goes uncovered.

To solve this problem, we propose Skill-Composer, an agent-native framework that performs the whole skill-selection process end to end. \textbf{Reframing the objective}: instead of ranking items by relevance, we select a non-redundant skill set over fragmented marketplaces, where each item is indispensable: dropping any one of them would leave a capability the task requires uncovered. \textbf{An end-to-end and agent-native design}: the framework carries out requirement clarification, cross-marketplace retrieval, and removing skills being capability overlap without any developer intervention. \textbf{Accountable coverage}: the decision moves from the developer to the agent, and the returned skill bundle carries justifications. This reframing changes the question that both marketplaces and retrieval research optimize, from ``which skills best match this query?'' to ``which minimal set of skills covers this task?''

\section{Motivation Example}\label{sec:motivation}

\begin{figure}[!ht]
\centering
\includegraphics[width=\columnwidth]{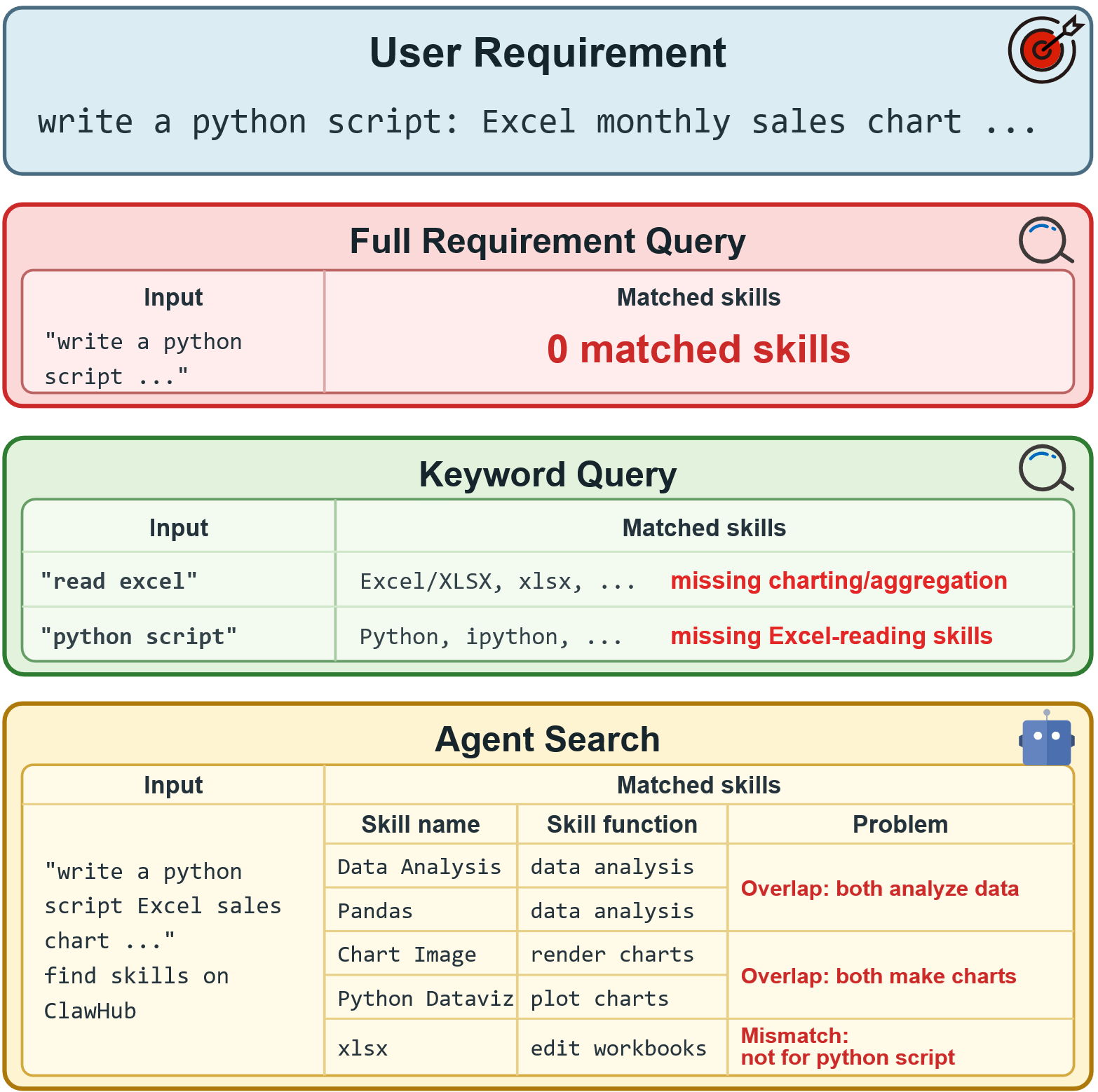}
\caption{A motivation example. A full-requirement query returns nothing, keyword queries yield separate skill lists needing to be cleared up, and agent search yields overlapping or mismatched skills.}
\label{fig:motivation}
\end{figure}

Consider the following request: \emph{Write a Python script that reads Excel sales records, aggregates sales by month, outputs a bar chart, and explains how to run it.}
Although concise, this requirement describes a workflow with several distinct obligations. Reading a workbook, cleaning amounts, aggregating records, generating a chart, and explaining execution require different capabilities. Further clarification also establishes requirements for invalid-value handling and readable Chinese chart labels.
We explore how ClawHub’s public API supports this task\footnote{\url{https://docs.openclaw.ai/clawhub/api.md}}.  Figure~\ref{fig:motivation} shows their results.


\subsection{Requirement and Keyword Querying}
This mirrors the situation a developer faces today: with a task stated in natural language, they open a skill marketplace and type it into the search box, retrieving the skills for the task. A \emph{requirement query} submits the task as one string and returns nothing, since the search only matches skills whose text aligns exactly with the query words. 
Splitting the requirement into eight \emph{keywords} as queries, such as \texttt{excel}, \texttt{data cleaning}, and \texttt{python script}, produced relevant candidates. However, the result lists do not explain how candidates complement one another or whether their combination covers the task. The caller still has to reconstruct the requirements and compare capabilities across lists.


\subsection{Agent Search}
A user can hardly search skills at scale or compare them one by one, and this is precisely what an agent is good at: it decides on its own what to look for and issues many queries. We therefore run an \emph{agent search}: Claude Code backed by DeepSeek-V4.1-Flash, to search ClawHub and recommend skills. However, each query returns a ranked list of redundant or incomplete skills. Even across three runs, the agent returned 11 distinct skills. One run returned five skills whose capabilities overlap in analyzing tables, and two for chart generation. Another recommendation, the anthropics xlsx skill, explicitly excluded tasks whose deliverable was a standalone Python script. Inspection also identified missing support for the Chinese labels required by the clarified task.



These observations motivate two selection criteria: coverage of the task’s requirements and a distinct contribution from each retained skill. Automated querying can reduce search effort, but it does not by itself enforce either criterion. The example therefore motivates an explicit procedure for assembling and explaining a bundle.


\section{Skill-Composer Framework}
\begin{figure*}[t]
\centering
\includegraphics[width=0.88\textwidth]{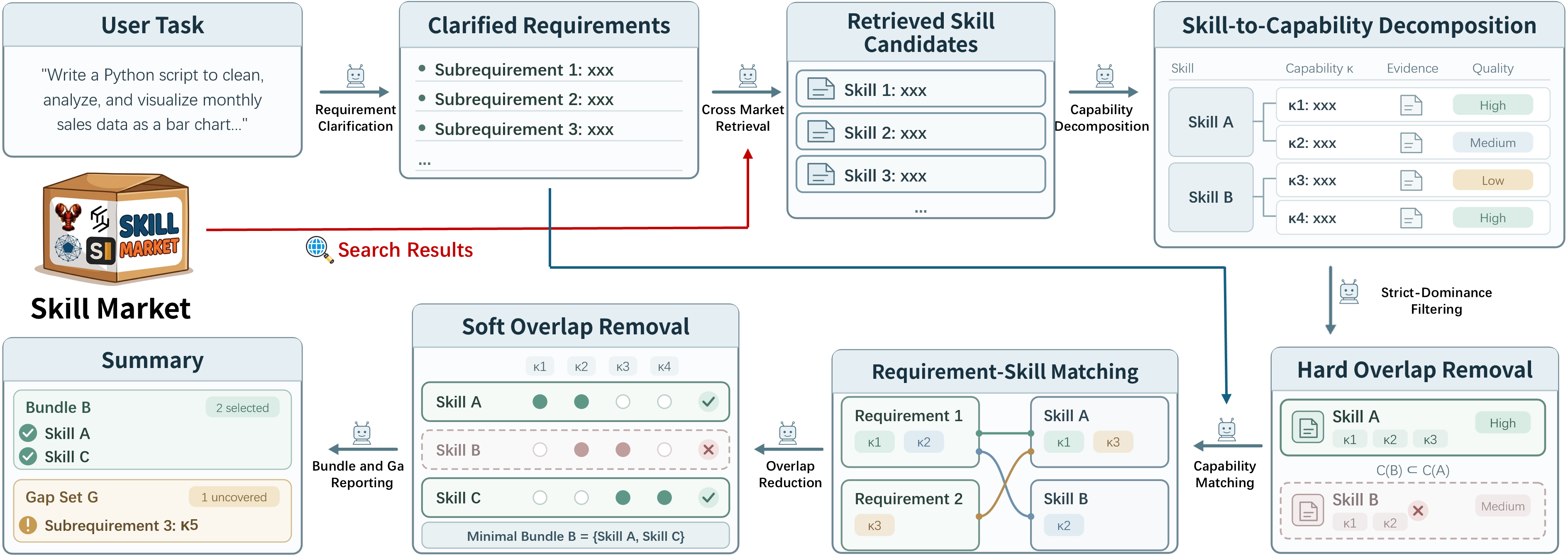}
\caption{Skill-Composer workflow: a task and marketplaces feed the agent, which applies six steps and returns a bundle $B$ and a summary.}
\label{fig:pipeline}
\end{figure*}

Skill-Composer takes a natural-language task and produces a skill bundle. It has six steps as shown in Figure~\ref{fig:pipeline}.

\textbf{Step 1: Requirement Clarification.} Given a task $T$, the agent clarifies its implicit constraints through targeted questions and decomposes it along functional boundaries into a requirement set $R=\{r_1,\dots,r_n\}$. Each requirement point is a tuple $r_i=\langle n_i,\sigma_i,\hat{C}_i\rangle$: an identifier $n_i$, a one-sentence statement $\sigma_i$ of what must be satisfied, and the set $\hat{C}_i$ of atomic capabilities that satisfy it. Requirement points and capability points are both expressed over a shared vocabulary $\mathcal{U}$ of atomic capabilities, so that the two sides can be compared directly.

\textbf{Step 2: Cross-Marketplace Skill Retrieval and Deduplication.} The agent builds queries from the requirement points and retrieves candidate skills across public marketplaces, including clawhub, skillhub, skillnet, and skillsmp. The hits are deduplicated: two candidates are judged identical when they match on their description, repository link, or body content, and only the most complete instance is kept. This yields the candidate pool $S=\{s_1,\dots,s_K\}$.

\textbf{Step 3: Skill Capability Decomposition.} Requirements and skills differ in granularity, so only by decomposing a skill into atomic capabilities can it be aligned with requirements and checked for overlap with others. Each skill $s\in S$ is decomposed into capability points, each a tuple $\langle\kappa,\varepsilon,q\rangle$ where $\kappa\in\mathcal{U}$ is an atomic capability the skill provides, $\varepsilon$ records the evidence in its body behind the quality judgment, and $q$ is that quality. We write $C(s)$ for the set of capabilities of $s$ and $q_s(\kappa)$ for its quality on $\kappa$, with $q_s(\kappa)\in\{\textit{lo},\textit{mid},\textit{hi}\}$. The three levels are totally ordered, so any two qualities are comparable. The evidence $\varepsilon$ serves auditability; it enters no rule.

\textbf{Step 4: Hard Overlap Removal.} Some candidates contain others, and retaining a subsumed skill adds no capability and only inflates the bundle, so hard overlap, that is, one capability set containing another, is removed first. We say $s$ dominates $t$, written $s\succeq t$, iff $C(t)\subseteq C(s)$ and $q_s(\kappa)\ge q_t(\kappa)$ for every $\kappa\in C(t)$. Since $\succeq$ is not antisymmetric, removing every dominated candidate would erase a whole class of mutually equivalent skills. Strict dominance, $s\succ t \iff s\succeq t \wedge \neg(t\succeq s)$, removes only the strictly dominated: $S'=\{t\in S\mid \nexists\,s\in S\setminus\{t\},\ s\succ t\}$. Mutually dominating skills are equivalent, and since neither is strictly dominated, a separate rule retains one representative of each equivalence class. The quality clause matters: a superset skill that is worse on the shared capabilities must not displace a better specialist.

\textbf{Step 5: Requirement-Skill Matching.} A skill $s$ contributes to requirement $r_i$ when $\hat{C}_i\cap C(s)\neq\emptyset$; satisfying $r_i$ requires all capabilities in $\hat{C}_i$. A candidate is dropped only if every required capability it provides is also provided by another retained skill; a sole provider of a required capability is never dropped, and when a match is unclear the candidate is kept, so that uncertainty favors coverage over parsimony. Dropping a candidate can make another a sole provider, so we repeat the rule until nothing changes and denote the resulting pool by $S''$.

\textbf{Step 6: Soft Overlap Removal.} The remaining candidates still share partial capabilities; keeping them one by one would provide the same capability repeatedly and leave the bundle non-minimal, so soft overlap, that is, partial intersection without containment, is not removed outright. Let $D=\bigcup_{i=1}^{n}\hat{C}_i$ and $D_{\mathrm{avail}}=D\cap\bigcup_{s\in S''}C(s)$. Among the remaining candidates we seek the smallest bundle covering the available required capabilities:
\begin{equation}
\min_{B\subseteq S''}|B|\quad\text{s.t.}\quad \bigcup_{s\in B}C(s)\supseteq D_{\mathrm{avail}} .
\label{eq:cover}
\end{equation}
We approximate \eqref{eq:cover} greedily; the minimum is not required exactly. Sole providers of required capabilities enter the bundle first; the greedy step then takes the skill covering the most not-yet-covered required capabilities, breaking ties by quality. A final pass then evicts any member whose required capabilities the rest already cover, restoring inclusion-wise minimality. The bundle covers all requirements only when $D_{\mathrm{avail}}=D$; otherwise it covers the available capabilities and reports the missing ones. Among bundles of equal size we prefer the one whose skills overlap least.

Finally, the agent returns not a ranked list but a bundle $B$ together with its gap set $G=\{r_i\mid \hat{C}_i\not\subseteq\bigcup_{s\in B}C(s)\}$, along with a concise summary of the selection.

\section{Case Study}
\begin{figure*}[t]
\centering
\includegraphics[width=0.88\textwidth]{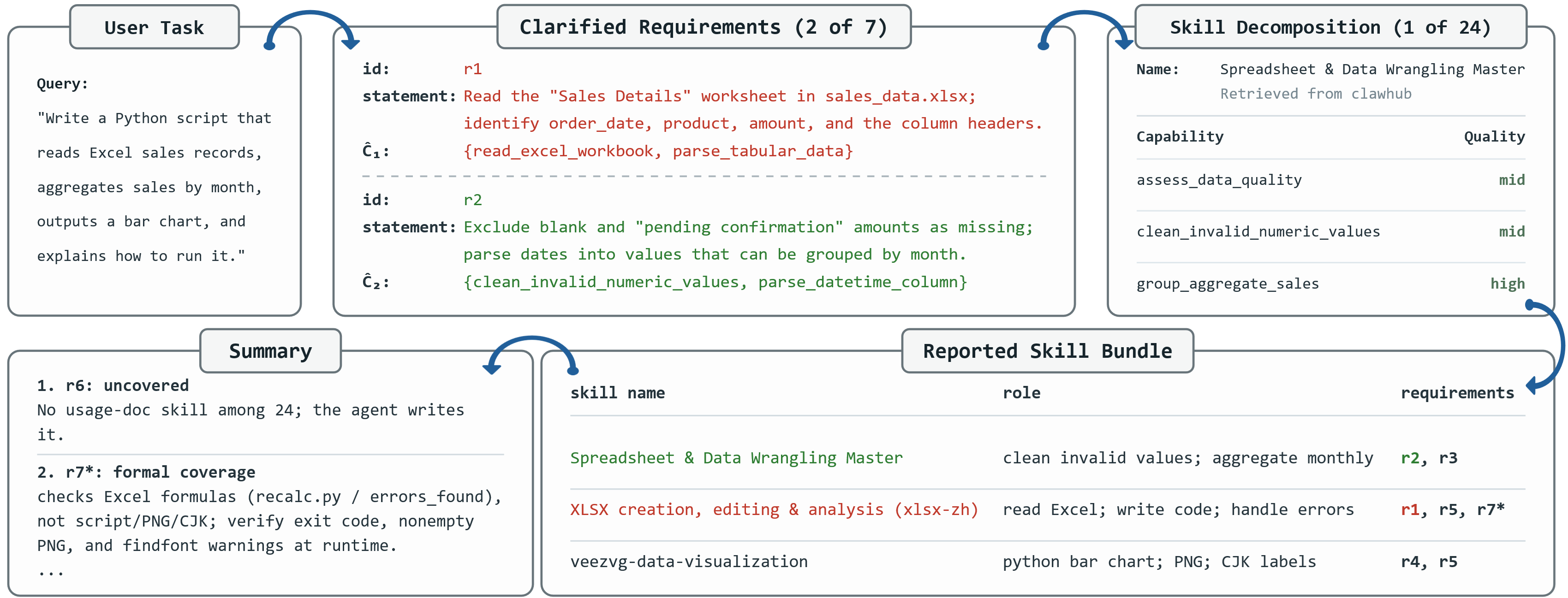}
\caption{Representative artifacts from the Python sales-analysis task: requirements, decomposition, bundle, and summary.}
\label{fig:case-study}
\end{figure*}

On the same task used in Section~\ref{sec:motivation}, we attempt to apply Skill-Composer. The results are shown in Figure~\ref{fig:case-study}. The code and data are provided at the link\footnote{\url{https://doi.org/10.6084/m9.figshare.35076466}}. The target data sit in a workbook of 32 rows, three of which carry a blank or a non-numeric amount, so the sentence hides a small data-cleaning obligation; space permits showing only two of the seven clarified requirements and one of the 24 retrieved candidates.

Clarification turns that sentence into seven requirement points, two of which are shown in the figure. Requirement $r_1$ pins down what sales records means: the worksheet, the three columns, the header row, and the capability pair \textit{read\_excel\_workbook} and \textit{parse\_tabular\_data}. Requirement $r_2$ supplies the adversarial reading of monthly aggregation: blank and pending amounts count as missing, and the date column must first be parsed into values that the aggregation can group by. Neither requirement appears in the sentence, and both produce the capabilities that later decide which skills enter the bundle.

Retrieval issues keywords derived from these points to four marketplaces and retrieves 24 candidate skills, which are then decomposed into capability points with quality grades. One candidate is shown in the figure. Its decomposition yields \textit{clean\_invalid\_numeric\_values} at mid quality, the capability $r_2$ needs, and \textit{group\_aggregate\_sales} at high quality, the capability $r_3$ needs.

After requirement matching and deduplication, selection returns three skills. \textit{Spreadsheet \& Data Wrangling Master} covers cleaning and monthly aggregation, \textit{xlsx-zh} covers reading Excel, writing the script, and handling errors, and \textit{veezvg-data-visualization} produces the bar chart as a PNG with readable CJK labels. Within the bundle, $r_4$ rests on \textit{veezvg-data-visualization} alone, while $r_5$ is served by both \textit{xlsx-zh} and \textit{veezvg-data-visualization} through different capabilities, so the final pass retains both.

The bundle does not cover every requirement: two of the seven are not truly satisfied. Requirement $r_6$, usage instructions, has no provider among the 24 candidates, so the agent reports it as the gap $G$ and writes the instructions itself rather than pad the bundle with a near miss. Requirement $r_7$ has formal coverage only: the corresponding \textit{xlsx-zh} verifies Excel formula recalculation rather than the running script or the generated PNG. Finally, the summary returns the recommended skill bundle and states which requirements remain uncovered. The case study thus traces the full path from a natural-language task to an installable skill bundle.

\section{Challenges and Opportunities}

\subsection{Requirements and Capability Granularity}
A bundle is only as good as the requirement set behind it. Clarification must add the implicit requirements a task needs to close, such as validation for a generated script, yet adding too much imports requirements the task never had; where to draw that line is a judgment we attempt to mitigate by showing the set to the user before retrieval. The same judgment governs decomposition: granularity must fall between ``process data'', too coarse to match any requirement, and ``call requests.get'', implementation detail, and because points are read off the body text alone, a briefly written skill is graded below what it does. Making the line and the split measurable, rather than a matter of discretion, is what we leave open.

\subsection{Minimality and Coverage Gaps}
Two limitations bound this step. First, finding the smallest bundle that covers everything available is a set cover problem, so we return a small bundle, not a provably smallest one. Second, minimization runs only over what the marketplaces supply: a capability no marketplace provides cannot be covered, and we report it as a gap rather than fill it with a near miss. A gap is informative because it has two possible causes: no skill covers that requirement, or one does but describes itself too vaguely to be decomposed. Separating the two would turn a gap from a failure into feedback that tells skill authors what to fix. This points to our most valuable opportunity, measurement: no benchmark yet connects a bundle to whether the task it was assembled for succeeds, so we cannot yet tell whether a reported gap reflects a real hole in the ecosystem or a weakness in our own decomposition.

\subsection{Trust and Provenance}
The current Skill-Composer procedure does not consider security at any step. The candidate skills come from different marketplaces, retrieval ranks them by relevance, and our comparison looks only at capability fit and how well a skill does its job, never at whether it can be trusted. A marketplace record does not say who published a skill, whether it is maintained, or what it may do once installed. Different marketplaces review their skills by their own rules, and a skill published in one is not necessarily reviewed by another, so a bundle that draws from several marketplaces can combine skills that were never checked under a common standard, and the user cannot see what was combined. This is a real gap in our work: cross-marketplace selection as we present it takes capability as the only criterion. Bringing provenance into selection, so that a bundle carries where each skill came from and how it was checked, is what we leave open.

\section{Conclusion}
This paper recasts skill acquisition from ranking to set selection: an agent decides which skills to combine, not which single skill looks closest to a query. For the agent, the choice of what to install passes from the developer to the agent itself; for marketplaces, capability labels matter more than keyword tags; for evaluation, skill quality should be read as bundle utility, not single-item relevance.

\bibliographystyle{IEEEtran}
\bibliography{references}

@misc{registry,
  author       = {Gao, Haoyu and Lulla, Jai Lal and Lin, Hong Yi and Baltes, Sebastian and
                  Treude, Christoph and
                  Zahedi, Mansooreh},
  title        = {{From Registry to Repository: How AI Agent Skills Are Written, Adapted, and Maintained}},
  type         = {arXiv preprint},
  archiveprefix = {arXiv},
  eprint       = {2607.00911},
  year         = {2026}
}

@misc{skillsmp,
  author       = {SkillsMP},
  title        = {{SkillsMP: Agent Skills Marketplace}},
  year         = {2026},
  howpublished = {\url{https://skillsmp.com}},
  note         = {Accessed: Nov. 2026}
}

@misc{clawhub,
  author       = {OpenClaw},
  title        = {{ClawHub Skills}},
  year         = {2026},
  howpublished = {\url{https://shturl.cc/FXsV8kTcvjn8xvv}},
  note         = {Accessed: Nov. 2026}
}

@misc{skillnet,
  author       = {Liang, Yuan and Zhong, Ruobin and Xu, Haoming and Jiang, Chen and
                  Zhong, Yi and Fang, Runnan and Gu, Jia-Chen and Deng, Shumin and
                  Yao, Yunzhi and Wang, Mengru and Qiao, Shuofei and Xue, Yida and Xu, Xin and
                  Wu, Tongtong and Wang, Kun and Liu, Yang and Bi, Zhen and Lou, Jungang and
                  Jiang, Yuchen Eleanor and Zhu, Hangcheng and Yu, Gang and Hong, Haiwen and
                  Huang, Longtao and Xue, Hui and Wang, Chenxi and Wang, Yijun and
                  Shan, Zifei and Chen, Xi and Tu, Zhaopeng and Xiong, Feiyu and Xie, Xin and
                  Zhang, Peng and Gui, Zhengke and Liang, Lei and Zhou, Jun and Wu, Chiyu and
                  Shang, Jin and Gong, Yu and Lin, Junyu and Xu, Changliang and
                  Deng, Hongjie and Zhang, Wen and Ding, Keyan and Zhang, Qiang and
                  Huang, Fei and Zhang, Ningyu and Pan, Jeff Z. and Qi, Guilin and
                  Wang, Haofen and
                  Chen, Huajun},
  title        = {{SkillNet}: Create, Evaluate, and Connect AI Skills},
  type         = {arXiv preprint},
  archiveprefix = {arXiv},
  eprint       = {2603.04448},
  year         = {2026}
}

@misc{toollery,
  author       = {Tian, Xiangxi and
                  Guan, Ran},
  title        = {{Toollery}: Scaling LLM Agents to Thousands of Skills and Tools},
  type         = {arXiv preprint},
  archiveprefix = {arXiv},
  eprint       = {2609.22218},
  year         = {2026}
}

@misc{swe-skills-bench,
  author       = {Han, Tingxu and Zhang, Yi and Song, Wei and Fang, Chunrong and
                  Chen, Zhenyu and Sun, Youcheng and
                  Hu, Lijie},
  title        = {{SWE-Skills-Bench}: Do Agent Skills Actually Help in Real-World Software Engineering?},
  type         = {arXiv preprint},
  archiveprefix = {arXiv},
  eprint       = {2603.15401},
  year         = {2026}
}

\end{document}